\documentclass[letterpaper, 10 pt, conference]{ieeeconf}  

\IEEEoverridecommandlockouts                              

\usepackage{graphicx}

\title{\LARGE \bf
Learning to Race in Minutes: Infoprop Dyna on the Mini Wheelbot}

\author{Devdutt Subhasish, Henrik Hose, and Sebastian Trimpe
\thanks{This work is funded in part by the German Research Foundation (DFG) under RTG 2236/2 (UnRAVeL) and the German Federal Ministry of Research, Technology and Space (BMFTR) under the Robotics Institute Germany (RIG), which the authors gratefully acknowledge. 
Further, the authors gratefully acknowledge the computing time provided to them at the NHR Center NHR4CES at RWTH Aachen University (project number p0022301).}
\thanks{Institute for Data Science in Mechanical Engineering, RWTH Aachen University, 52062 Aachen, {\small\tt \{devdutt.subhasish, henrik.hose, trimpe\}@dsme.rwth-aachen.de}}
}

\begin{document}

\maketitle
\thispagestyle{empty}
\pagestyle{empty}

\begin{abstract}

Reinforcement Learning (RL) has the potential to enable robots with fast, nonlinear, and unstable dynamics to reach the limits of their performance. 
However, most recent advances rely on carefully designed physics-based simulators and domain randomization to achieve successful sim-to-real transfer within reasonable wall-clock time. 
In this work, we bypass the need for such simulators and demonstrate that Infoprop Dyna, a state-of-the-art uncertainty-aware model-based reinforcement learning (MBRL) framework, can enable robots to learn directly from real-world interactions. 
Using Infoprop Dyna, the Mini Wheelbot, an underactuated unicycle robot, learns to race around a track within 11 minutes of real-world experience.

\end{abstract}

\section{Introduction}
Learning-based control of robotic systems with fast, unstable, and nonlinear dynamics remains challenging, particularly when learning directly on real hardware. 
Reinforcement learning (RL) has shown strong potential in this domain \cite{kaufmann_champion-level_2023}, but conventional RL methods suffer from poor sample efficiency, limiting their practical applicability. 
As a result, many successful approaches rely on carefully designed physics-based simulators and extensive domain randomization to enable sim-to-real transfer of the learned policy within reasonable wall-clock time \cite{openai_solving_2019,rudin_learning_2022, smith_walk_2022}.

Model-based reinforcement learning (MBRL) addresses this limitation by learning a model of the system dynamics and using it to augment real experience with imagined rollouts \cite{deisenroth_pilco_2011, hafner_dream_2020, frauenknecht_trust_2024}. 
While this can significantly improve sample efficiency, imperfectly learned models can lead to model exploitation, especially when used for long-horizon planning, resulting in degraded real-world performance. 
Infoprop Dyna is a recent MBRL method that mitigates this issue, resulting in reliable rollouts over longer horizons than any prior method \cite{frauenknecht_rollouts_2025}.
Treating the model predictions as noisy observations of a ground truth signal, Infoprop Dyna leverages information theory to recover the estimated ground truth signal, while simultaneously keeping track of the accumulated corruption due to noise.
Infoprop Dyna has shown state-of-the-art performance on a suite of simulated MuJoCo tasks \cite{todorov_mujoco_2012}.

In this work, we demonstrate the practical effectiveness of Infoprop Dyna on a challenging real-world control task. 
We consider the Mini Wheelbot \cite{hose_mini_2025}, an underactuated unicycle robot with highly nonlinear yaw dynamics, and task it with learning to race around a fixed track.
Prior to this work, no controller had been designed for this task.
Using Infoprop Dyna, the agent learns directly from real-world interactions, progressing from conservative motion to aggressive racing behavior within minutes and discovering non-trivial strategies such as controlled slipping at high speeds.

\begin{figure}
    \centering
    \includegraphics[width=0.71\linewidth]{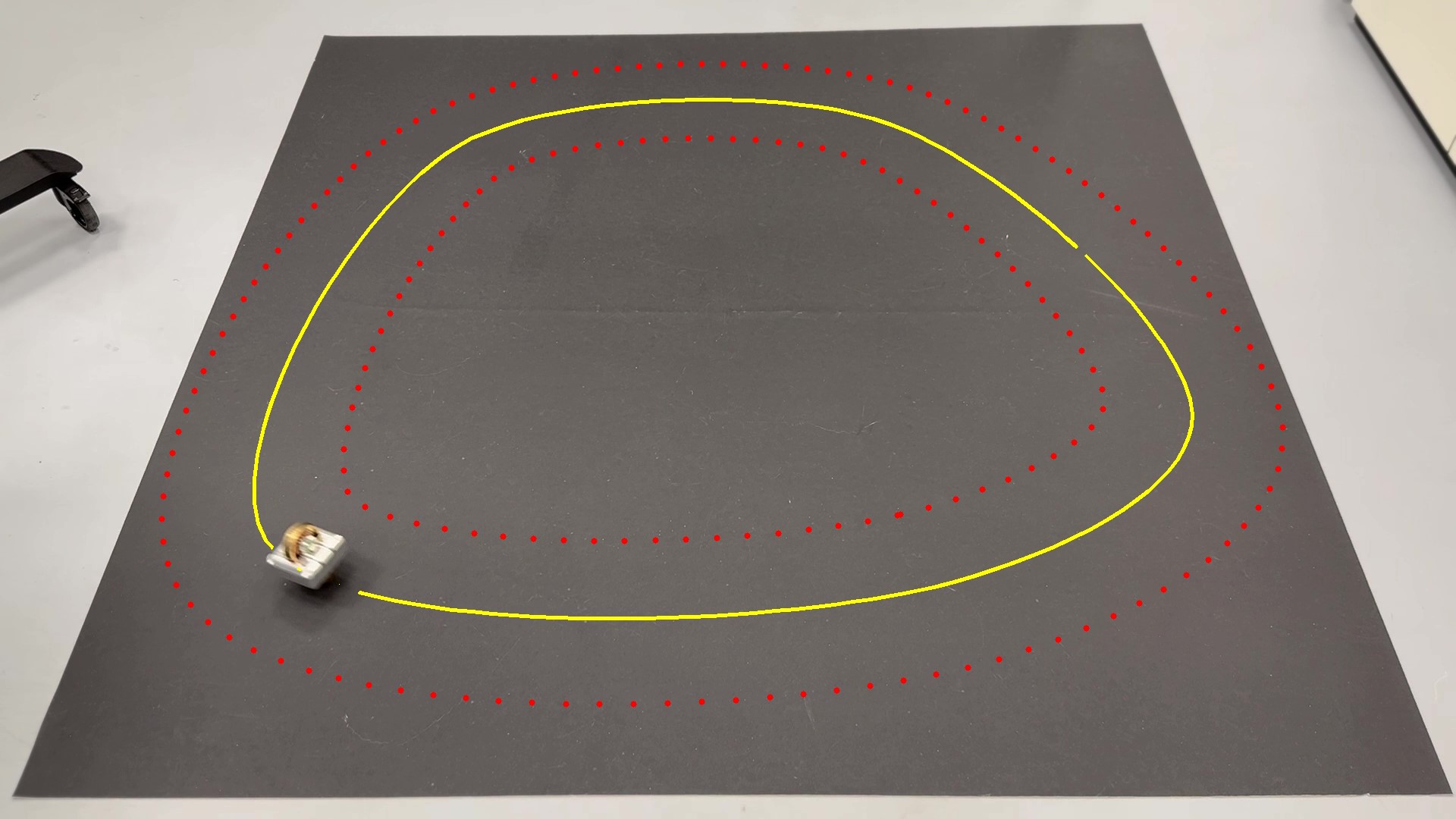}
    \caption{The Mini Wheelbot racing arena.}
    \label{fig:arena}
\end{figure}

\begin{figure}
    \centering
    \includegraphics[width=0.91\linewidth]{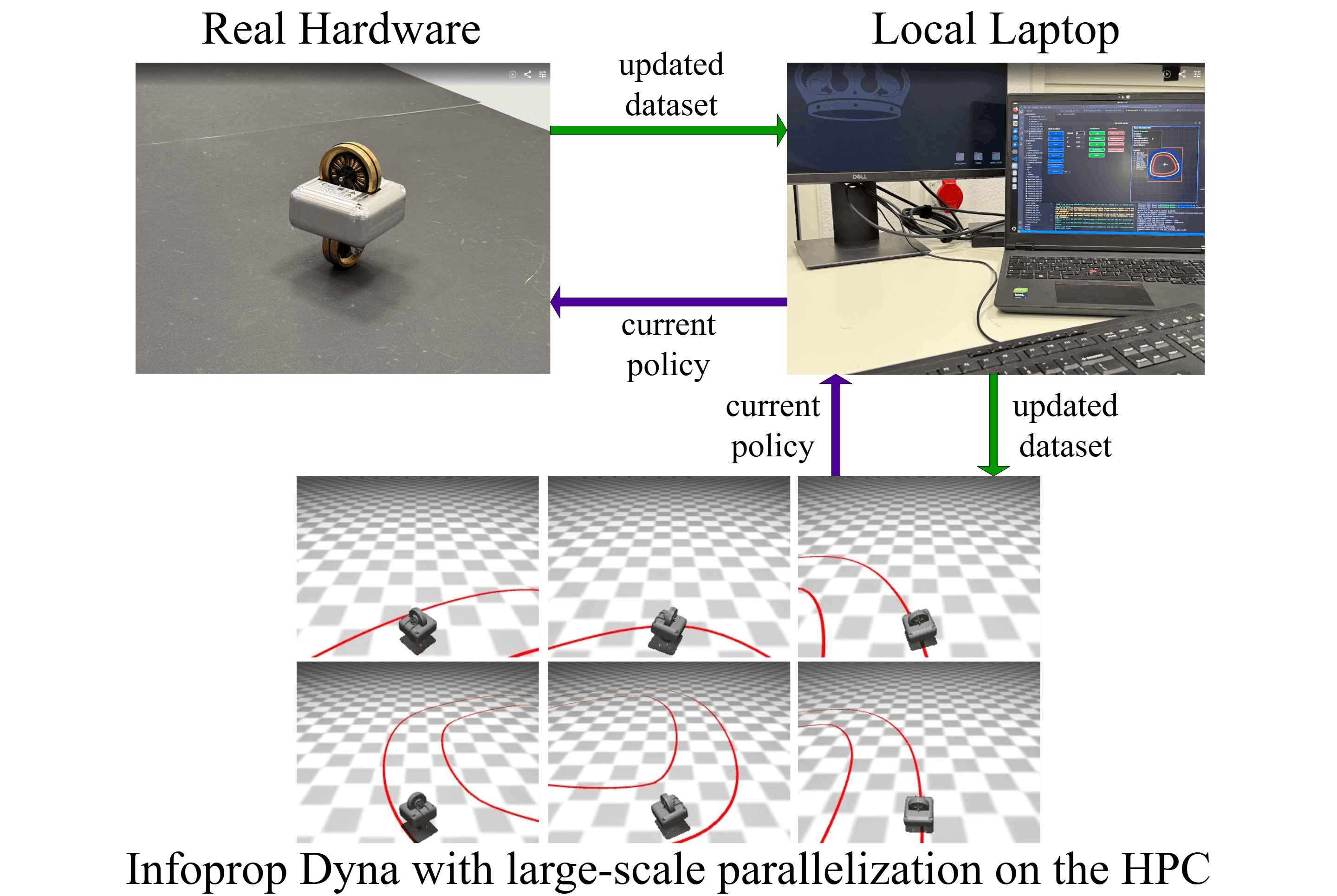}
    \caption{Distributed training schematic. A learned dynamics model generates large-scale parallel synthetic rollouts, replacing a physics-based simulator.}
    \label{fig:pipeline}
\end{figure}

\section{Challenges of Mini Wheelbot Racing}

The Mini Wheelbot presents a challenging control problem due to its yaw degree of freedom \cite{hose_mini_2025}.
The nonlinear yaw dynamics is indirectly influenced by wheel actuation via gyroscopic effects and geometric turning.
It is strongly dependent on ground contact interactions. 
This necessitates longer horizon planning for effective trajectory tracking behavior.
In addition, the compact form factor of the Mini Wheelbot leads to small time constants in its dominant modes of motion. 
Small differences in actuation can rapidly lead to qualitatively different system behaviors. 
This sensitivity is further amplified during high-speed motion, where contact-induced nonlinearities and slip effects become significant and difficult to model accurately.
The combination of difficult to model dynamics and need for long-horizon rollouts make Infoprop Dyna a natural fit for the Mini Wheelbot racing task.

\section{Experimental Setup}

\textbf{Task.}
The objective is to learn a racing policy that drives the Mini Wheelbot around a fixed track, shown in Figure \ref{fig:arena}, as fast as possible. 
Rather than formally defining the task in terms of state-action spaces and reward, we provide a high-level description. 
The agent observes the Wheelbot’s physics state, as defined in the Mini Wheelbot paper \cite{hose_mini_2025}, together with a local representation of the track given by the relative positions of the next 30 track points expressed in polar coordinates. 
The racing agent directly commands the motor torques. 
The reward encourages faster velocities along the track, penalizes deviation from the track centerline, and assigns a terminal penalty for crashes or leaving the track.

\textbf{Dynamics model learning.}
We only learn a transition model of the Wheelbot’s physics state. 
The relative track representation, is deterministically reconstructed from the predicted physics state and the known track geometry. 
In contrast to prior Infoprop benchmarks \cite{frauenknecht_rollouts_2025}, which assume access to fully observed, memoryless states, the Mini Wheelbot only provides estimated states obtained from onboard state estimation \cite{hose_mini_2025}. 
To account for this, the learned model is conditioned on a short history of estimated physics states and applied control inputs and is trained to predict the next estimated state. 
This allows the model to implicitly capture the dynamics of the state estimator alongside the physical system.

\textbf{Training pipeline.}
Training is performed in a distributed setup, illustrated in Figure \ref{fig:pipeline}. 
Real-world data is collected by executing the current policy on the physical robot. 
Logged trajectories are transmitted via SSH to a local workstation, which handles bookkeeping. 
The dataset is then forwarded to a high-performance computing (HPC) cluster, where model learning and policy optimization are performed using a JAX-based \cite{noauthor_quickstart_nodate} implementation built on top of BRAX \cite{noauthor_googlebrax_2026}. 
Note that we do not use any simulation model throughout the training.
Trained policy parameters are periodically saved, transferred back to the local workstation, and deployed on the robot for the next round of data collection and evaluation. 
This pipeline enables training times that are comparable to real-world data collection times through massive parallelization on the HPC.
This is in contrast to the original Infoprop benchmarks where training dominated wall-clock time.

\textbf{Initialization.}
The real-world dataset is warm-started using trajectories collected by manually driving the robot around the track with an approximate model predictive controller (AMPC) \cite{hose_mini_2025} and a joystick interface. 
This provides an initial one minute of safe and informative transitions before fully autonomous learning begins.

\section{Training Run}

We now describe a single continuous real-world training run of the Mini Wheelbot racing agent. 
This section can be best understood using the accompanying video for visual reference. 
The training starts with initial data collection using the joystick-driven AMPC. 
Following this, the agent gradually learns to turn and negotiate corners within the first five minutes of real world experience.
After experiencing six minutes of real interactions, the agent successfully completes its first lap of the track. 
Over the subsequent minutes, lap completion becomes increasingly consistent. 
By seven to eight minutes, the agent completes multiple laps, although some turns are still executed conservatively and noticeable wobbling remains during high-curvature segments.
Starting from nine minutes, the agent masters the track, achieving peak performance after 11 minutes of real world experience.

\begin{figure}
    \centering
    \includegraphics[width=0.91\linewidth]{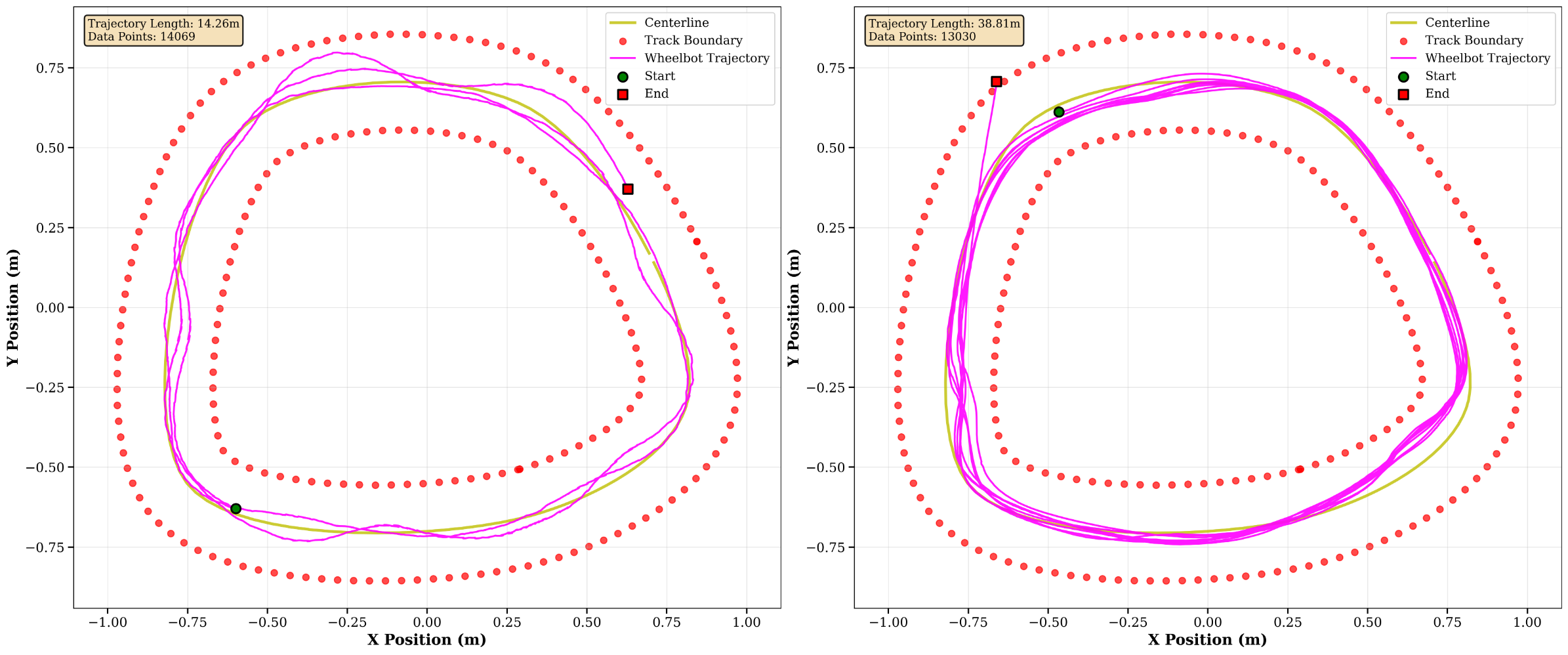}
    \caption{Trajectory from the AMPC (left) and final racing agent (right).}
    \label{fig:comparison}
\end{figure}

Figure \ref{fig:comparison} compares the trajectory generated by the AMPC~\cite{hose_mini_2025} and the final racing agent. 
Apart from sticking more closely to the track centerline, the racing agent completes more than three times the number of laps as the AMPC agent. 
The AMPC has an average speed of $0.15\,\mathrm{m/s}$ and a maximum speed of $0.33\,\mathrm{m/s}$.
The final racing agent reaches an average speed of $0.5\,\mathrm{m/s}$ and a peak speed of $0.97\,\mathrm{m/s}$. 
To achieve this speed-up, the racing agent exploits a controlled slipping behavior to handle fast corners. 
This is unlike the AMPC, which uses gyroscopic effects to reorient itself. 
This behavior is non trivial to capture using purely physics-based simulation. 

\section{Conclusion and Next Steps}

We present a real-world demonstration of rapid, high-performance learning on a challenging robotic platform using Infoprop Dyna. 
The Mini Wheelbot acquires aggressive racing behavior within minutes, exhibiting non-trivial behaviors, such as controlled slipping, without relying on a carefully engineered sim-to-real pipeline. 
These results highlight the practical potential of Infoprop Dyna to bridge the gap between data efficiency and high final performance on real robotic systems.

Looking forward, several directions can further strengthen and extend this demonstration. 
We plan to scale the setup to larger arenas and arbitrary track layouts, enabling higher speeds and requiring greater generalization across trajectories. 
We also want to develop a more general and modular JAX-based distributed Infoprop Dyna interface that can be readily applied to a broader range of robotic systems. 
The hope is that these steps will facilitate wider adoption of real-world model-based reinforcement learning within the robotics community.

\bibliographystyle{IEEEtran}
\bibliography{references}
\end{document}